\documentclass[runningheads]{llncs}
\usepackage[T1]{fontenc}
\usepackage{graphicx,verbatim}

\usepackage{hyperref}
\usepackage{amssymb}
\usepackage[normalem]{ulem}
\useunder{\uline}{\ul}{}
\usepackage{multirow}
\usepackage[dvipsnames]{xcolor}
\usepackage[misc]{ifsym}

\usepackage{color}

\begin{document}
\title{Learning Contrastive Multimodal Fusion\\with Improved Modality Dropout\\for Disease Detection and Prediction}
\titlerunning{Learning Multimodal Fusion}
%
\author{Yi Gu\inst{1,2} \and
Kuniaki Saito\inst{1} \and
Jiaxin Ma\inst{1}\textsuperscript{(\href{mailto:jiaxin.ma@sinicx.com}{\Letter)}}
}
%
%
\authorrunning{Y. Gu et al.}
%
\institute{OMRON SINIC X Corporation\\
\email{jiaxin.ma@sinicx.com}\and
Nara Institute of Science and Technology
}
\maketitle              
\begin{abstract}
As medical diagnoses increasingly leverage multimodal data, machine learning models are expected to effectively fuse heterogeneous information while remaining robust to missing modalities.
In this work, we propose a novel multimodal learning framework that integrates enhanced modalities dropout and contrastive learning to address real-world limitations such as modality imbalance and missingness.
Our approach introduces learnable modality tokens for improving missingness-aware fusion of modalities and augments conventional unimodal contrastive objectives with fused multimodal representations.
We validate our framework on large-scale clinical datasets for disease detection and prediction tasks, encompassing both visual and tabular modalities.
Experimental results demonstrate that our method achieves state-of-the-art performance, particularly in challenging and practical scenarios where only a single modality is available.
Furthermore, we show its adaptability through successful integration with a recent CT foundation model.
Our findings highlight the effectiveness, efficiency, and generalizability of our approach for multimodal learning, offering a scalable, low-cost solution with significant potential for real-world clinical applications.
The code is available at \href{https://github.com/omron-sinicx/medical-modality-dropout}{https://github.com/omron-sinicx/medical-modality-dropout}.


\end{abstract}
\section{Introduction}
Advancements in medical diagnostic have led to increasingly diverse clinical data, comprising multimodal sources such as medical images (e.g., computed tomography [CT] and magnetic resonance imaging) and tabular data (e.g., electronic health records [EHRs] and radiology reports).
This diversity has driven research in multimodal learning to enhance predictive modeling in health care \cite{chen_unified_2024,hager_best_2023,polsterl_combining_2021,robinet_drim_2024,ding_hia_2024,gao_medbind_2024,jiang_mgdr_2024,xiong_mome_2024,liu_multi-modal_2024,qiu_multimodal_2022,xiong_multi-modality_2024,zhou_pathm3_2024,grzeszczyk_tabattention_2023,feng_unified_2024,zhang_m2fusion_2024,jain_mmbcd_2024,xu_temporal_2024}.
Deep learning models trained on multimodal data have demonstrated improved performance in disease detection and prediction, including lung diseases analysis \cite{kim_llm-guided_2024,huang_multimodal_2020,zhou_radfusion_2021}.
Despite these advancements, effectively leveraging multimodal information remains challenging due to real-world constraints such as modality imbalance and missingness.

Previous worsk have attempted to mitigate these challenges through modality dropout, wich enhances model robustness by simulating missing modalities during training \cite{hussen_abdelaziz_modality_2020,neverova_moddrop_2016,krishna_modality_2024,qi_multimodal_2024}.
However, the traditional modality dropout uses fixed placeholders, limiting its ability to improve missingness awareness.
Recent methods have explored learnable missingness instructions \cite{chen_unified_2024,xu_temporal_2024,lee_multimodal_2023}; however, their integration into modality dropout remains underexplored.
On the other hand, contrastive learning has emerged as a powerful technique for multimodal representation learning \cite{hager_best_2023,robinet_drim_2024,zhang_m2fusion_2024,gao_medbind_2024,jiang_mgdr_2024,xiong_multi-modality_2024,radford_learning_2021,kim_learning_2021,zhai_sigmoid_2023,yu_coca_2022}.
By encouraging models to associate information from heterogeneous sources that refer to the same underlying concept (e.g., the same patient or event), contrastive learning improved downstream task performance.
However, most contrastive methods focus on unimodal representations, whereas the fused multimodal representations were not utilized.

In this research, we propose a novel multimodal learning framework designed to enhance both unimodal and multimodal performance for disease detection and prediction.
We build a multimodal model consisting of unimodal encoders, a neural fusion module, and a task-specific head.
We assume the unimodal encoders are pretrained and frozen during our training to demonstrate the improvement with a low cost of additional training.
Unlike previous work that produced unimodal representations by unimodal encoders, our method leverages the multimodal model to generate unimodal representations with modality dropout.
Additionally, we introduce learnable modality tokens in modality dropout to improve the model's awareness of missing modalities.
Furthermore, we propose multimodal contrastive learning with fused multimodal representations for better representation binding.
We validate our method using two large-scale public clinical datasets with three tasks of disease detection and prediction from CT and tabular data.
We also employ a recent CT foundation model \cite{yang_advancing_2024} as the encoder and show the improvement by our method.
\textbf{The contribution} of this work is threefold:
1) We propose a novel multimodal learning framework with improved modality dropout and contrastive multimodal learning.
2) We demonstrate the effectiveness and efficacy of the proposed method using large-scale public clinical datasets for disease detection and prediction.
3) We show the efficient improvement on a recent CT foundation model at a low cost.

\section{Method}
\begin{figure}
\includegraphics[width=\textwidth]{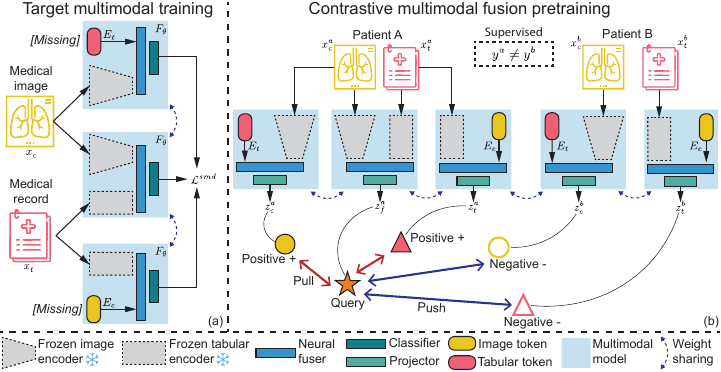}
\caption{
Overview of the proposed method for training the multimodal model.
(a) Target training using the simultaneous modality dropout, where the multimodal model is supervised using unimodal and multimodal inputs simultaneously.
(b) Inter-modality contrastive learning using the multimodal and unimodal representations.
The learnable modality tokens are introduced for improving missingness-aware neural fusion.
} \label{fig:overview}
\end{figure}

The proposed multimodal learning framework is illustrated in Fig. \ref{fig:overview}.
We aim to train a multimodal model $F_\theta(\cdot,\cdot)$ (parameterized by $\theta$) that detects or predicts diseases from medical images and tabular data while being robust to missing modalities during inference.
Our model comprises pretrained unimodal encoders for processing individual modalities, a lightweight fusion module of multilayer perception (MLP) to integrate information from different sources, and a task-specific head, which is a classifier for the target training (Fig. \ref{fig:overview} [a]) or a projector for pretraining (Fig. \ref{fig:overview} [b]).
The unimodal encoders remain frozen throughout all training to reduce additional training overhead.

\subsubsection{Simultaneous modality dropout.}
We consider the task of detecting or predicting disease from medical images and tabular data as illustrated in Fig. \ref{fig:overview} (a).
Given a patient sample, with image $x^i_c$ and tabular $x^i_t$ modalities, the multimodal model $F_\theta(\cdot,\cdot)$ predicts the probability $p(y^i\mid x^i_c,x^i_t,\theta)=F_\theta(x^i_c,x^i_t)$, where $y^i$ is the associated label.
The model is trained by maximizing the log-likelihood using $\mathcal{L}^{base}=-\mathrm{log}p(y^i\mid x^i_c,x^i_t,\theta)$.
We adopt modality dropout \cite{hussen_abdelaziz_modality_2020,neverova_moddrop_2016,krishna_modality_2024,qi_multimodal_2024} to enable robustness to missing modalities.
Unimodal predictions, $p(y^i\mid x^i_c,\theta)=F_\theta(x^i_c,\textbf{0}_t)$ and $p(y^i\mid x^i_t,\theta)=F_\theta(\textbf{0}_c,x^i_t,)$, are obtained by replacing missing modalities with zero matrices, where the $\textbf{0}_c$ and $\textbf{0}_t$ are the zero metrics for the image and tabular modalities, respectively.
Traditional modality dropout training $\mathcal{L}^{md}=-\mathrm{log}p(y^i\mid\{x^i_j\}_{j\in S},\theta)$ employs a random sampling function $g(\cdot)$ to select a subset $S=\{g(M)\in\mathcal{P}(M)\setminus \varnothing\}$ of available modalities at each iteration, where $\mathcal{P}(M)$ represents the power of all modalities $M=\{c,t\}$.
The sampling function $g(\cdot)$ was introduced to avoid large computational costs in a single iteration as the number of modality combinations $|\mathcal{P}(M)|=2^{|M|}$ scales exponentially \cite{neverova_moddrop_2016}.
Instead, we propose simultaneous modality dropout, where all modality combinations are explicitly supervised without sampling, leveraging the small number of modalities and the lightweight nature of our fusion module.
Our loss function is defined as Eq. \ref{eq:smd}, where the $\lambda$ is a hyperparameter that balances term weights.
This way, our approach ensures a smoother loss gradient, leading to more stable training.
Since the unimodal encoders remain frozen, we apply modality dropout at the input of the fusion module to prevent redundant computation.

\begin{equation}
\label{eq:smd}
    \mathcal{L}^{smd}= -\mathrm{log}p(y^i\mid,x^i_c,x^i_t,\theta)
     -\lambda\sum_{j\in M}\mathrm{log}p(y^i\mid x^i_j,\theta)
\end{equation}

\subsubsection{Learnable modality tokens.}
While conventional modality dropout effectively handled missing modalities during inference, its reliance on fixed zero matrices ($\textbf{0}_c$ and $\textbf{0}_t$) limited the performance of multimodal learning.
Inspired by recent multimodal methods that employed learnable instructions \cite{chen_unified_2024,xu_temporal_2024,lee_multimodal_2023}, we introduce learnable modality tokens $E_c$ and $E_t$ to replace fixed zero matrices in our modality dropout strategy.
Specifically, the unimodal predictions of the multimodal model $F_\theta(\cdot,\cdot)$, originally defined as $F_\theta(x^i_c,\textbf{0}_t)$ and $F_\theta(\textbf{0}_c,x^i_t,)$, are replaced with $F_\theta(x^i_c,E_t)$ and $F_\theta(E_c,x^i_t,)$, respectively.
This adaptation enhances the model's generalization of missing modalities while preserving representational consistency.
The modality tokens are integrated into the inputs of neural fusion modules for efficiency (shown in Fig. \ref{fig:overview}).

\subsubsection{Contrastive multimodal fusion.}
Conventional contrastive learning primarily focused on unimodal representations to enhance the performance of downstream tasks \cite{qu_multi-modal_2024,robinet_drim_2024,kim_learning_2021,hager_best_2023,zhai_sigmoid_2023}.
In contrast, our method incorporates fused multimodal representations into contrastive learning, facilitating better cross-modal representation binding (see Fig. \ref{fig:overview} [b]) to improve both unimodal and multimodal performance.
Following \cite{zhai_sigmoid_2023}, we employ the sigmoid-based contrastive loss for computational efficiency.
We adopt supervised contrastive learning, leveraging label information, as it has been shown to outperform self-supervised approaches \cite{zhai_sigmoid_2023,khosla_supervised_2020}.
Let $z^i_c$, $z^i_t$, and $z^i_f$ denote the encoded representations for the image, tabular, and image-tabular data of the $i$-th patient, respectively.
For a batch of $n$ patients with indices $N=\{1,2,...,n\}$, we define the contrastive loss between modalities $i$ and $j$ as Eq. \ref{eq:con},
where the $a(u,v)=1$ for positive pair ($y^u=y^v$) and $a(u,v)=-1$ for negative pair ($y^u\neq y^v$).
Our approach builds upon conventional contrastive learning, initially defined as $\mathcal{L}^{con}=\mathcal{L}^{con}_{c,t}$.
Unlike softmax-based contrastive learning, which requires separate directional losses between modalities (i.e., $\mathcal{L}^{con}_{c,t}\neq \mathcal{L}^{con}_{t,c}$), simoid-based contrastive learning is undirectional, meaning $\mathcal{L}^{con}_{c,t}=\mathcal{L}^{con}_{t,c}$.
We extend contrastive learning to fused representations, incorporating multimodal information into the alignment process.
Our final contrastive loss is defined as $\hat{\mathcal{L}}^{con}=\mathcal{L}^{con}_{c,t}+\mathcal{L}^{con}_{c,f}+\mathcal{L}^{con}_{t,f}$.
This way, the multimodal representation $z^i_f$ serves as a robust intermediary between unimodal ones $z^i_c$ and $z^i_t$ to improve representation alignment and more effective multimodal fusion.
\begin{equation}
\label{eq:con}
    \mathcal{L}^{con}_{i,j}=-\sum_{u\in N}\sum_{v\in N}\mathrm{log}\frac{1}{1+e^{a(u,v)(-tz^u_i\cdot z^v_j+b)}},
\end{equation}

\section{Experiment}
To validate our proposed method, we conducted experiments on two publicly available dtasets: the Multimodal Pulmonary Embolism (PE) dataset \cite{zhou_radfusion_2021,huang_penetscalable_2020} and the National Lung Screening Trial (NLST) dataset \cite{nlst_2013}.
The PE dataset consists of 1,837 chest CT scans from 1,794 patients, with EHRs containing PE diagnosis results.
Among these CT scans, 1,111 (60.48\%) are labeled as PE-positive.
Following prior work \cite{zhou_radfusion_2021,huang_penetscalable_2020}, we formulated the PE detection task as a binary classification problem, utilizing both CT image and tabular data.
The NLST dataset comprises 64,117 chest CT scans from 12,498 patients, each associated with EHRs.
The objective, as outlined in the CT foundation model demo \cite{yang_advancing_2024}, is to predict future cancer occurrence at one-year and two-year intervals, treating these as independent classification tasks.
The dataset includes 868 (6.95\%) and 1,438 (11.51\%) CT scans labeled as cancer-positive within one and two years, respectively.
Our experimental setup aligns with these objectives, evaluating model performance in PE detection and cancer prediction by integrating multimodal learning techniques.

\subsubsection{Experimental setting.}
We employed a four-fold cross-validation strategy at the patient level for both datasets.
Within each fold, we further reserved 10\% of the training data for validation.
To benchmark our approach, we compared it against recent unimodal and multimodal baselines.
Additionally, we conducted ablation studies to systematically evaluate the contributions of individual components of our proposal.
Since our tasks were essentially doing classification, we utilized two categories of classification evaluation metrics: 1) Probability-estimation metrics (assess model confidence without thresholding), including area under the receiver operating characteristic curve (AUROC), average precision (AP), and the area under the risk-coverage curve (AURC). 2) decision-making metrics (evaluate binary classification with predefined threshold, e.g., 0.5), including Matthews correlation coefficient (MCC) and the F-score.
We report the best and second-best results using bold and underlined fonts, respectively.
During training, we used the AUROC from validation data to capture the best model, as it has been shown to be robust to class imbalance \cite{mcdermott_closer_2024}
For the PE detection task, we report all the evaluation metrics.
In contrast, for cancer prediction tasks using NLST dataset, we focus on probability-estimation metrics for clarity and consistency, given the high sensitivity of decision-based metrics to threshold selection in extremely class-imbalanced settings.

\subsubsection{Implementation details.}

For PE detection, we utilized two unimodal encoders, PENet \cite{huang_penetscalable_2020} (for CT) and FT-Transformer \cite{gorishniy_revisiting_2021} (for tabular data), followed by a lightweight one-layer MLP (for multimodal fusion) and a linear regression head, which predicted label probabilities.
To reduce the computational overhead, we froze the unimodal encoders during training, updating only the MLP and head, resulting in an efficient model with only 2.52 M \textit{additional} training parameters.
The CT images were resized to $256\times256$; the training batch size was set to 8.
In prior PE dataset studies \cite{huang_penetscalable_2020,zhou_radfusion_2021},  models were trained using 24-slice CT windows due to computational constraints.
At inference, the patient-level predictions were aggregated using maximum pooling across CT windows.
However, we found that feeding the entire CT directly improved inference performance, so we adopted full-image training and inference.
For the cancer prediction tasks, we employed the CT foundation model \cite{yang_advancing_2024} as our image encoder.
When we implemented the CT foundation model as an image-only baseline method, we followed the demo to train a two-layer MLP to predict the labels using embeddings extracted by the CT foundation model.
The training batch size was set to 128.
For all tasks, we used SHAP \cite{lundberg_unified_2017} to select the most related attributes in tabular data.
After tuning, top-8 and top-16 attributes were selected for the PE and NLST datasets, respectively.
We used AdamW \cite{loshchilov_decoupled_2018} optimizer with $1\times10^{-4}$ learning rate and $1\times10^{-4}$ weight decay to train our method for 150 epochs.
We simply used $\lambda=1$ for all experiments, as it demonstrated robustness within the range 0.5 to 2.0.
A fixed random seed was carefully maintained throughout training to ensure a fair comparison across experiments.

\begin{table}
\caption{Result of PE dataset}\label{tab:pe}
\centering
\begin{tabular}{lccccccccccccc}
\hline
\multirow{2}{*}{} & \multirow{2}{*}{} & \multicolumn{3}{c}{Inference} & \multirow{2}{*}{} & \multirow{2}{*}{AUROC$\uparrow$} & \multirow{2}{*}{} & \multirow{2}{*}{AP$\uparrow$} & \multirow{2}{*}{} & \multirow{2}{*}{AURC$\downarrow$} & \multirow{2}{*}{} & \multirow{2}{*}{MCC$\uparrow$} & \multirow{2}{*}{F-score$\uparrow$} \\ \cline{3-5}
 &  & CT &  & Table &  &  &  &  &  &  &  &  &  \\ \hline
PENet \cite{huang_penetscalable_2020}&  & \checkmark &  &  & {\ul } & 0.758 &  & 0.609 &  & 0.475 &  & 0.207 & 0.538 \\
PENet $\dagger$ \cite{huang_penetscalable_2020} &  &  &  &  &  & {\ul 0.778} &  & {\ul 0.680} &  & {\ul 0.442} &  & {\ul 0.379} & {\ul 0.614} \\
Ours $\dagger$ &  &  &  &  &  & \textbf{0.801} &  & \textbf{0.724} &  & \textbf{0.422} &  & \textbf{0.451} & \textbf{0.647} \\ \hline
ElasticNet \cite{zou_regularization_2005}&  &  &  & \checkmark &  & \textbf{0.758} &  & \textbf{0.581} &  & \textbf{0.487} &  & \textbf{0.370} & 0.564 \\
FT-Transformer \cite{gorishniy_revisiting_2021}&  &  &  &  &  & 0.745 &  & 0.539 &  & 0.510 &  & 0.351 & \textbf{0.597} \\
Ours $\dagger$ &  &  &  &  &  & {\ul 0.751} &  & {\ul 0.558} &  & {\ul 0.499} &  & {\ul 0.352} & {\ul 0.594} \\ \hline
DAFT \cite{polsterl_combining_2021}&  & \checkmark &  & \checkmark &  & 0.739 &  & 0.536 &  & 0.511 &  & 0.334 & 0.595 \\
DAFT $\dagger$ \cite{polsterl_combining_2021}&  &  &  &  &  & 0.629 &  & 0.459 &  & 0.566 &  & 0.174 & 0.437 \\
DAFT-64 \cite{polsterl_combining_2021}&  &  &  &  &  & 0.700 &  & 0.459 &  & 0.566 &  & 0.259 & 0.561 \\
DAFT-64 $\dagger$ \cite{polsterl_combining_2021}&  &  &  &  &  & 0.616 &  & 0.479 &  & 0.560 &  & 0.139 & 0.434 \\
TabAttention \cite{grzeszczyk_tabattention_2023}&  &  &  &  &  & 0.738 &  & 0.551 &  & 0.505 &  & 0.271 & 0.565 \\
RadFusion \cite{zhou_radfusion_2021}&  &  &  &  &  & 0.811 &  & 0.676 &  & 0.438 &  & 0.294 & 0.572 \\
RadFusion $\dagger$ \cite{zhou_radfusion_2021}&  &  &  &  &  & {\ul 0.819} &  & {\ul 0.716} &  & 0.422 &  & 0.418 & 0.633 \\
RadFusion (FT) &  &  &  &  &  & 0.803 &  & 0.642 &  & 0.454 &  & 0.288 & 0.567 \\
RadFusion (FT) $\dagger$&  &  &  &  &  & 0.815 &  & 0.707 &  & {\ul 0.426} &  & {\ul 0.425} & {\ul 0.640} \\
Ours $\dagger$&  &  &  &  &  & \textbf{0.842} &  & \textbf{0.775} &  & \textbf{0.397} &  & \textbf{0.499} & \textbf{0.676} \\ \hline
\end{tabular}
\end{table}

\subsubsection{PE dataset results.}
Table \ref{tab:pe} represents the performance comparison between our method and existing approaches including image-only models (PENet \cite{huang_penetscalable_2020}), tabular-only models (ElasticNet \cite{zou_regularization_2005} and FT-Transformer \cite{gorishniy_revisiting_2021}), and the multimodal models (DAFT \cite{polsterl_combining_2021}, TabAttention \cite{grzeszczyk_tabattention_2023}, and RadFusion \cite{zhou_radfusion_2021}).
We categorize results based on inference settings: image-only, tabular-only, and image-tabular.
Our method was trained using both image and tabular data while also being robust to modality missingness during inference.
The results show that our approach outperformed conventional image-only and tabular-only methods.
Compared to PENnet, we increased the AUROC and AP from 0.758 to 0.801 and 0.609 to 0.724, respectively.
While ElasticNet exhibited the best tabular-only performance due to the limited number of training samples, our method achieved the best performance in the image-tabular settings.
RadFusion, in its original implementation, utilized PENet and ElasticNet.
To ensure a fair comparison, we also evaluated a RadFusion variant that replaced ElastiveNet with FT-Transformer (denoted as RadFusion [FT]).
Our method, with only a few additional training parameters, outperformed Radfusion, improving image-tabular AUROC from 0.803 to 0.842 and demonstrating superior multimodal fusion efficiency and robustness.

\begin{table}
\caption{Result of NLST dataset}\label{tab:nlst}
\centering
\begin{tabular}{llclclclclclc}
\hline
 &  & \multicolumn{5}{c}{Cancer in 2 years} &  & \multicolumn{5}{c}{Cancer in 1 year} \\ \cline{3-7} \cline{9-13} 
 &  & AUROC$\uparrow$ &  & AP$\uparrow$ &  & AURC$\downarrow$ &  & AUROC$\uparrow$ &  & AP$\uparrow$ &  & AURC$\downarrow$ \\ \hline
CT foundation model \cite{yang_advancing_2024}&  & 0.729 &  & 0.070 &  & 0.956 &  & 0.700 &  & 0.050 &  & 0.972 \\
Ours (CT only) &  & 0.732 &  & 0.068 &  & 0.956 &  & 0.780 &  & 0.068 &  & 0.969 \\
ElasticNet \cite{zou_regularization_2005}&  & 0.808 &  & 0.199 &  & 0.938 &  & 0.830 &  & 0.132 &  & 0.962 \\
FT-Transformer \cite{gorishniy_revisiting_2021}&  & {\ul 0.837} &  & {\ul 0.246} &  & {\ul 0.933} &  & {\ul 0.925} &  & {\ul \textbf{0.279}} &  & {\ul \textbf{0.949}} \\
Ours &  & \textbf{0.857} &  & \textbf{0.247} &  & \textbf{0.931} &  & \textbf{0.926} &  & {\ul \textbf{0.279}} &  & {\ul \textbf{0.949}} \\ \hline
\end{tabular}
\end{table}

\subsubsection{NLST dataset results.}
Table \ref{tab:nlst} represents the evaluation results on the NLST dataset, comparing our method against the CT foundation model \cite{yang_advancing_2024}, ElasticNet \cite{zou_regularization_2005}, and FT-Transformer \cite{gorishniy_revisiting_2021}.
We report our model's performance under image-only (denoted as Ours [CT-only]) and image-tabular inference settings.
Our method outperformed the CT foundation model in both two-year and one-year cancer prediction tasks, improving AUROC from 0.729 to 0.732 and 0.700 to 0.780, respectively.
When incorporating tabular data, our model achieved the best overall performance, demonstrating the benefit of our multimodal learning.
Interestingly, we observed that tabular data dominated the prediction in this dataset, as the performance gap between our image-tabular model and the tabular-only model was minimal.
This suggested that morphometric attributes in the tabular data provide highly discriminative features for cancer prediction.
Despite this, our approach successfully integrated CT imaging and tabular data, offering the best overall performance while improving image-only one at a low additional training cost.
this observation aligns with prior work \cite{hager_best_2023}, which demonstrated the effectiveness of morphometric attributes in contrastive learning for medical analysis.

\subsubsection{Ablation study.}
To assess the contribution of individual components in our proposed method, we conducted ablation studies shown in Table \ref{tab:abl}.
For simplicity, we report only AUROC, as it provides a comprehensive evaluation across all tasks.
We evaluated our method across image-only and image-tabular inference settings.
We denote the one-year and two-year cancer prediction tasks as NLST$_1$ and NLST$_2$, respectively.
We first train the model using only the standard cross-entropy loss $\mathcal{L}^{base}$, establishing a baseline.
The proposed simultaneous modality dropout $\mathcal{L}^{smd}$ slightly improved the conventional one $\mathcal{L}^{md}$.
The AUROC values were improved from 0.836 to 0.840 and 0.712 to 0.722 for image-tabular PE detection and image-only NLST$_2$, respectively.
Integrating learnable modality tokens further enhanced the performance for both $\mathcal{L}^{md}$ and $\mathcal{L}^{smd}$.
Adding conventional contrastive learning $\mathcal{L}^{con}$ before target training with $\mathcal{L}^{smd}$ and learnable modality tokens further improved the performance.
Finally, replacing $\mathcal{L}^{con}$ with the proposed contrastive learning $\hat{\mathcal{L}}^{con}$ consistently achieved the best performance across all tasks.
These results validated the effectiveness of the proposed simultaneous modality dropout, learnable modality tokens, and contrastive multimodal fusion, demonstrating their collective contribution to robust multimodal learning.

\begin{table}
\caption{Ablation study result (AUROC$\uparrow$)}\label{tab:abl}
\centering
\begin{tabular}{ccccccccccccccccc}
\hline
\multirow{2}{*}{\begin{tabular}[c]{@{}c@{}}Training\\ loss\end{tabular}} &  & \multirow{2}{*}{\begin{tabular}[c]{@{}c@{}}Modality\\ token\end{tabular}} &  & \multirow{2}{*}{\begin{tabular}[c]{@{}c@{}}Pretraining\\ loss\end{tabular}} &  & \multicolumn{5}{c}{Image-tabular inference} &  & \multicolumn{5}{c}{Image-only inference} \\ \cline{7-11} \cline{13-17} 
 &  &  &  &  &  & PE &  & NLST$_2$ &  & NLST$_1$ &  & PE &  & NLST$_2$ &  & NLST$_1$ \\ \hline
$\mathcal{L}^{base}$ &  & / &  & / &  & 0.837 &  & 0.847 &  & 0.919 &  & / &  & / &  & / \\
$\mathcal{L}^{md}$ &  & - &  &  &  & 0.836 &  & 0.850 &  & 0.917 &  & 0.797 &  & 0.714 &  & 0.707 \\
 &  & \checkmark &  &  &  & 0.840 &  & 0.855 &  & 0.915 &  & 0.796 &  & {\ul 0.728} &  & 0.741 \\
$\mathcal{L}^{smd}$ &  & - &  &  &  & 0.838 &  & 0.851 &  & {\ul \textbf{0.926}} &  & 0.800 &  & 0.716 &  & 0.738 \\
 &  & \checkmark &  &  &  & 0.840 &  & 0.855 &  & 0.920 &  & {\ul 0.800} &  & 0.722 &  & 0.751 \\
 &  &  &  & $\mathcal{L}^{con}$ &  & {\ul \textbf{0.842}} &  & {\ul 0.856} &  & {\ul \textbf{0.926}} &  & {\ul 0.800} &  & 0.726 &  & {\ul 0.771} \\
 &  &  &  & $\hat{\mathcal{L}}^{con}$ &  & {\ul \textbf{0.842}} &  & \textbf{0.857} &  & {\ul \textbf{0.926}} &  & \textbf{0.801} &  & \textbf{0.732} &  & \textbf{0.780} \\ \hline
\end{tabular}
\end{table}

\section{Conclusion}
We introduced a novel multimodal learning framework that integrates modality dropout with contrastive multimodal pertaining to enhance disease detection and prediction from CT images and tabular data.
Our approach incorporated learnable modality tokens to improve missing modality awareness and leveraged fused multimodal representations in contrastive learning for improved alignment across modality representations.
Through evaluations on three multimodal prediction tasks from two datasets, we demonstrated the effectiveness of our method in both unimodal and multimodal inference settings, showcasing its practical applicability.
Using frozen unimodal encoders, we achieved performance gains with a minimal cost of \textit{additional} training.
Further improvements are anticipated if these encoders are unfrozen and trained end-to-end.
Our framework is scalable to additional modalities beyond CT and EHR data.
Future work will explore this scalability by validating performance across a broader range of modalities and investigating its applicability in conjunction with large language models, particularly the major decoder-only architectures, across various clinical tasks, advancing the potential of multimodal learning in clinical applications.

%
%
%
\bibliographystyle{splncs04}
\bibliography{mybibliography_short}
%




\end{document}